%% file: main.tex
\documentclass[11pt]{article}

\usepackage[english]{babel}

\usepackage[letterpaper,top=2cm,bottom=2cm,left=3cm,right=3cm,marginparwidth=1.75cm]{geometry}

\usepackage{amsmath}
\usepackage{graphicx}
\usepackage{booktabs}
\usepackage{array}
\usepackage{placeins} 
\usepackage[colorlinks=true, allcolors=blue]{hyperref}
\usepackage[round]{natbib}
\usepackage{chngcntr}
\usepackage{appendix}
\usepackage{longtable}
\usepackage{tabularx}
\usepackage{relsize} 

\title{\textbf{Assessing The Potential Of Mid-Sized Language Models For Clinical QA}}
\author{Elliot Bolton¹*\textsuperscript{\dag}, Betty Xiong¹*, \and Vijaytha Muralidharan¹, Joel Schamroth², Vivek Muralidharan³, \and Christopher D. Manning¹**\textsuperscript{\dag}, Roxana Daneshjou¹**  \\[1ex] 
¹Stanford University \\ 
²University College London \\
³The University of Cambridge \\[1ex]
\textsuperscript{\dag}Correspondence: elliotbolton@stanford.edu, manning@stanford.edu \\ * Denotes equal contribution\hspace{1em}** Denotes equal PI contribution}

\date{April 2024}

\begin{document}
\maketitle
\begin{abstract}
\noindent Large language models, such as GPT-4 and Med-PaLM, have shown impressive performance on clinical tasks; however, they require access to compute, are closed-source, and cannot be deployed on device. Mid-size models such as BioGPT-large, BioMedLM, LLaMA 2, and Mistral 7B avoid these drawbacks, but their capacity for clinical tasks has been understudied. To help assess their potential for clinical use and help researchers decide which model they should use, we compare their performance on two clinical question-answering (QA) tasks: MedQA and consumer query answering. We find that Mistral 7B is the best performing model, winning on all benchmarks and outperforming models trained specifically for the biomedical domain. While Mistral 7B’s MedQA score of 63.0\% approaches the original Med-PaLM, and it often can produce plausible responses to consumer health queries, room for improvement still exists. This study provides the first head-to-head assessment of open source mid-sized models on clinical tasks.
\end{abstract}

\section{Introduction}

Recently, large language models (LLMs) models, such as GPT-4 \citep{openai2024gpt4} and Med-PaLM~2 \citep{singhal2023expertlevel}, have achieved impressive performance in clinical question-answer (QA) tasks, with a GPT-4 based system achieving 90.2\% on the MedQA task \citep{nori2023generalist} and Med-PaLM~2 producing responses to consumer health questions competitive with human physicians. However, there are multiple drawbacks to these models. Their parameter counts can range into the trillions that require dedicated compute clusters, making them expensive to train, expensive to run, and environmentally unsustainable. These massive models are closed off from researchers, only accessible via a paid API. This means researchers and practitioners cannot study these models and research on improvements are limited to those with access to the model weights and architecture. The closed nature of these models requires users to communicate with them via the internet. Applications involving language model analysis of sensitive patient info would need to be sent to a third party, raising serious HIPAA compliance issues.

A new paradigm, on-device AI (or edge AI), involves running language models on a local device such as a phone or tablet. This could have many uses in biomedicine, providing medical knowledge during natural disasters or in remote locations where internet access is poor or non-existent. The closed nature and size of models such as GPT-4 and Med-PaLM~2 makes them unsuitable for on-device AI.

Open source, mid-size language models (\textless 10B parameters) can address these shortcomings. They offer cost-effective and environmentally friendly alternatives. They can be downloaded and used on organization’s internal clusters, their architectures and parameters are freely available, and their reasonable size means they can plausibly be run on portable devices.

There are two categories of model relevant to the biomedical setting. Smaller domain-specific models (\textless 3B parameters) such as BioGPT-large \citep{Luo_2022} and BioMedLM \citep{bolton2024biomedlm} were trained exclusively on biomedical text from PubMed. Larger 7B parameter models such as LLaMA 2 \citep{touvron2023llama} and Mistral 7B \citep{jiang2023mistral} are more powerful, but were trained on general English text and lack the biomedical focus of their smaller counterparts.

It is an open question which model is most suitable for a clinical QA application, and what level of performance can be achieved with these models. Does training exclusively on biomedical text offer clear performance gains over training on general English data?

To help address these questions, we rigorously test all 4 models in the clinical QA domain on 2 popular tasks which test the ability to comprehend and reason about medical scenarios and produce informative paragraph responses to health questions: MedQA (USMLE-style questions) and MultiMedQA Long Form Answering (open response to consumer health queries).

\section{Methods}

\subsection{MedQA}

\subsubsection{MedQA task description}

The MedQA four-option task \citep{jin2020disease} involves answering a standard USMLE-style question with four multiple-choice options. This task has become a standard benchmark used to evaluate language model’s capacities for utilizing medical knowledge and reasoning about clinical scenarios. Questions can range from requesting specific medical knowledge (e.g., symptoms of schizophrenia) to the presentation of a clinical scenario and a request for the most appropriate diagnosis or course of action (e.g., A 27-year-old male presents …). The official test set contains 1273 questions for system evaluation.

\subsubsection{MedQA fine tuning}

The MedQA dataset comes with 10178 training examples, 1272 development examples, and 1273 test examples. Every example was divided into a prompt and expected response. The format is presented in the supplementary material.

All four models were provided the same prompt and trained to generate the response, which simply consisted of the text ``Answer: '' and the letter of the correct option.
\paragraph{4-Way Model Comparison} All four models were fully fine-tuned on the 10178 training examples, meaning all of their parameters were updated. To ensure a fair comparison between the models, the same format, training data, and training code was used for all models. Models were fine-tuned with the Hugging Face library.

It is important to note that the same set of hyperparameters will not necessarily be ideal for two different models. Furthermore, models of different sizes require different ranges of learning rates. It has been shown larger models benefit from smaller learning rates. To ensure a fair comparison, it is important to run a similar hyperparameter sweep for each model with appropriate values.

For each hyperparameter setting, each model was fine-tuned three times with three different random seeds, and an average score on the development set was calculated. The hyperparameter setting with the highest average score on the development set was chosen for each model, and those hyperparameters were used to evaluate the model on the test set.

\paragraph{Additional Mistral 7B Experiments} To further explore the capabilities of mid-size models, the model with the best performance (Mistral 7B) was fine-tuned on the concatenation of the MedQA training data and the larger MedMCQA training set which contains 182822 additional examples. Training on this data has been shown to boost MedQA performance \citep{bolton2024biomedlm, singhal2023expertlevel}. A slightly more elaborate prompt was used for this phase, and the model was trained to generate both the correct letter and the full text of the answer. A similar hyperparameter sweep was employed to determine the best settings. It is important to note these experiments were focused on maximizing Mistral 7B performance, not on producing a fair comparison to other models.

Details of the hyperparameter sweeps and prompts used can be found in part A and B of the supplementary material.

\subsection{MultiMedQA Long Form Question Answering}

\subsubsection{MultiMedQA Long Form Question Answering task description}

The MultiMedQA Long Form Question Answering task \citep{singhal2022large} involves presenting the model with consumer health questions typical of those issued to search engines. 4000 questions come from three datasets: LiveQA, MedicationQA, and HealthSearchQA. LiveQA additionally comes with reference answers. The system is expected to generate a comprehensive answer of one or two paragraphs on par with a health FAQ page response.

The questions cover a wide array of consumer health topics, including symptoms and treatments of various conditions, infectious diseases, chronic illnesses, nutritional deficiencies, reproductive health, developmental disorders, medication usage, medication interactions, and preventative measures.

\subsubsection{MultiMedQA Long Form Question fine tuning}

There is no standard, publicly available training data for this task, so a curated training set was created from a mixture of publicly available data sources on the web. Pages containing information about medical topics were uniquely parsed to produce question and response pairs.

Some pages had an FAQ-style format. For example, over 3500 entries from the MedLinePlus Encyclopedia were used. The page was parsed, and question, response examples were extracted.

Other sources did not have explicit question, response examples, but questions were derived from section headings. For instance, Wikipedia pages were a rich source of examples. Sections were extracted from Wikipedia pages on medical topics, the section header was translated into a question format, and the corresponding section text was used as the response to the question. Common sections found in Wikipedia included ``Contraindications'', ``Side Effects'', ``Signs and symptoms'', ``Risk factors'', and ``Treatment''. These section titles were translated into question format.

As a specific example, consider the Wikipedia page for ``Arthritis''. The page topic was extracted as arthritis. A section on this page entitled ``Treatment'' was mapped to the template ``What are the treatments for \_\_ ?''. Combining the page topic with the template produced the question ``What are the treatments for arthritis?''. The section text was used as the response to this question.

Overall 61,400 training examples and 1000 development examples were produced. These examples were divided into prompt, response format, and all 4 models were trained to generate the response given the prompt with the Hugging Face library.

The hyperparameter sweep process was similar to the one employed for MedQA. Hyperparameters were chosen based on their performance on the development set. The metric used to assess performance was the model loss on the entire response text given the prompt.

\subsubsection{Clinician Review Of Responses To Health Questions}

The physicians reviewed the responses to the questions along multiple axes (see table~\ref{tab:rubric}).

\begin{table*}[p]
    \centering
    \small
    \begin{tabular}{p{3.25cm} p{11.5cm}}
    \toprule
    Dimension & Description and Likert Scale \\
    \midrule
    \\
    Complete & {The extent to which necessary content to answer the question is presented

    \bigskip
    5 excellent, covers all major relevant points needed to answer the question
    
    4 good, covers most relevant points, missing a key detail or two
    
    3 moderate, covers some important information but is also missing significant details
    
    2 poor, at best contains one correct fact related to the question
    
    1 zero correct information relevant to the question is presented
    
    } \\
    
    Error Free & {The extent to which no medical errors are present in the response

    \bigskip
    5 zero medical errors
    
    4 one medical error
    
    3 two medical errors
    
    2 three medical errors
    
    1 4+ medical errors

    } \\
    
    Appropriate & {The extent to which the answer is free of extraneous content?

    \bigskip
    5 excellent, appropriate interpretation and all necessary assumptions stated
    
    4 good, appropriate interpretation but some assumptions not stated
    
    3 moderate, somewhat appropriate interpretation but missing assumptions
    
    2 poor, interpretation of question is flawed but has some relevance
    
    1 zero attempt at answering the question appropriately

    } \\
    
    Harm Extent & {The potential harm the response could cause?
    
    \bigskip
    5 no harm
    
    4 mild harm
    
    3 moderate harm
    
    2 severe harm
    
    1 death or life-threatening injury
    
    } \\
    
    Harm Likelihood & {The likelihood the answer will cause harm

    \bigskip
    5 Very Unlikely
    
    4 Unlikely
    
    3 Maybe
    
    2 Likely
    
    1 Certainly
    
    } \\
    
    Bias & {The extent to which the response is biased against a subgroup

    \bigskip
    5 no bias
    
    4 mild bias
    
    3 moderate bias
    
    2 considerable bias
    
    1 severe bias
    
    } \\
    \bottomrule
    \end{tabular}
    \caption{Clinical Review Rubric For Medical Question Responses}
    \label{tab:rubric}
\end{table*}

Although the Med-PaLM paper introduced interesting axes for assessment, we found it to be lacking in nuance. Its dimensions across comprehension, retrieval and reasoning were highly correlated, i.e., a language model has to do a combination of all of the above to provide a response, and it is difficult to disentangle one dimension from the other. It only allowed for binary indicators, i.e., most of the questions were binary and instantiated with the question: ``Does the answer contain any evidence of correct / incorrect ...?'' In this case, it is impossible to differentiate whether a minor error of retrieval was found, or whether the entire generated response was rife with errors.

With extensive consultation and input from four physicians, we streamlined the twelve Med-PaLM dimensions to six in our rubric. To introduce more nuance in each indicator, we developed a Likert scale from one to five, as shown in previous studies \citep{Tang2023.04.22.23288967, sullivan2013likert}. With the Likert scale, we can better quantify values such as how complete an answer is or how many errors it contains. For each dimension, we provide a detailed description of what each point on the Likert scale should correspond to, which harmonizes the overall format of the clinician grading.

A high quality response to a medical query should be \textit{complete}: provide essential, correct information to allow a user to address a medical need. It should be comprehensive, communicating all necessary information to address the question. It must be \textit{error-free}: devoid of medical errors or risk causing confusion and harm. It should be \textit{appropriate}: given the context of the question, the answer should triage the situation and address it accordingly, as would a trained physician. Its \textit{harm extent} (the potential for harm) \textit{harm-likelihood} (likelihood to cause harm) should be minimized, and it should not perpetuate \textit{bias} (against a subgroup). The dimensions of our rubric correspond to these aspects of a high quality answer.

We further segmented the types of questions into categories, with the options of Prognostic, Treatment, Diagnosis, Severity, Risk factor and Other. We borrowed concepts from previous studies on assessing human evaluation of text generation \citep{smith2022human, celikyilmaz2021evaluation, see2019makes}, especially in expert-related tasks \citep{malaviya2024expertqa}. This allows us to ascertain error rates for different types of questions and perform better statistical analyses.

To assess the quality of the generated answers, we included three physician reviewers. They were tasked to independently assess outputs generated by the models. Initially, there was a training session where the physicians discussed a subset of questions and agreed upon a set of standards moving forward in the grading. Following on, each reviewer was asked to review 45 questions, each with four different generations from the different models. The order of the question-answer pairs was randomized, the model that generated the answer was blinded from the reviewers, the formatting was uniform across all models, and the reviewers were explicitly instructed to assess the generation as a standalone. These measures were designed to minimize potential bias in grading that arise due to order or model effects.

The significance of 5-point Likert scales between models was calculated by the Mann-Whitney U test \citep{nachar2008utest}. Furthermore, the analysis was rerun with the stratification of the different question categories. The numerical scores for the response categories of a 5-point Likert item range from one to five. These scores were utilized in the Mann-Whitney U test to assess differences. The $p$-value indicates whether there is a distinction in the responses of summaries produced by the two models. This is based on the assumption that the null hypothesis posits no variance between the results generated by the two models. The mean is reported for each of the models across the different criteria dimensions.

\subsection{Mid-Size Model Details}

It is fruitful to review key details of the four models (see table \ref{tab:models}). This can help practitioners decide which model is best for their use case.

We provide a summary of each model considering the following aspects:

\textbf{Model size}, which influences expense, speed, and potential for on-device usage.

\textbf{Training dataset}, the data on which the model is trained on, if known.

\textbf{Context length}, i.e., the maximum number of tokens the model can use to influence its prediction of the next token. Note for generative tasks, the prompt and desired response must both fit in the context length for tokens of the response to be influenced by the prompt. This is particularly relevant in the retrieval augmented generation scenario (R.A.G.).

\textbf{Position encoding}, i.e., how the model encodes position. Models with learned embeddings can only be finetuned to perform tasks with the same context length as they were trained with during pretraining, since they have learned embeddings for each context position. Models that utilize RoPE can be extended at finetune time to potentially work with longer contexts due to RoPE’s flexibility.

\begin{table}[ht]
    \centering
    \label{tab:models_summary}
    \begin{tabular}{lllll}
    \toprule
    Model & Size & Training Set & Context & Position Encoding \\
    \midrule
    BioGPT-large & 1.5B & PubMed Abstracts & 1024 & Learned embedding \\
    BioMedLM & 2.7B & PubMed Abstracts, PubMed Central & 1024 & Learned embedding \\
    LLaMA 2 7B & 7B & Unknown & 4096 & RoPE \\
    Mistral 7B & 7B & Unknown & 8000 & RoPE \\
    \bottomrule
    \end{tabular}
    \caption{Summary of Various Models}
    \label{tab:models}
\end{table}
\FloatBarrier

\section{Results}

We evaluated four open source models from the sub-10B parameter class: BioGPT-large, BioMedLM, LLaMA 7B and Mistral 7B.

\subsection{MedQA Performance Results}

These four models yielded the following results (see table \ref{tab:medqa}).

\begin{table}[ht]
    \centering
    \begin{tabular}{l r l r}
     \toprule
     Model & Params & Accuracy \\
     \midrule
     BioGPT-large & 1.5B & 40.7 \\
     BioMedLM & 2.7B & 46.3 \\
     LLaMA 2 & 7B & 47.3 \\
     Mistral 7B & 7B & 59.1 \\
     \midrule
     \textbf{Mistral 7B (w/MedMCQA data)} & \textbf{7B} & \textbf{63.0} \\
     \bottomrule
    \end{tabular}
    \caption{MedQA Performance Of Models After Instruction Tuning
    }
    \label{tab:medqa}
\end{table}
\FloatBarrier

The strongest performer was Mistral 7B (59.14\%), with a clear performance difference of 11.88\% in performance compared to the next best model, LLaMA 2 7B (47.26\%).

We took the strongest performer Mistral 7B and trained it on the combination of the MedQA and MedMCQA training data sets (193k examples) and further boosted performance, achieving a score of 63.0\%.

\subsection{Clinician Review Of MultiMedQA Long Form Question Answering Results}

The physicians gave the four models the following average scores across the six dimensions (see table \ref{tab:medmcqa}).

\begin{table}[ht]
    \centering
    \begin{tabular}{l l l l l l l}
     \toprule
     Model & Complete & Error Free & Appropriate & Harm  & Harm & Bias \\
      & & & & Extent & Likelihood & \\
     \midrule
     BioGPT-large & 3.72 & 3.97 & 3.91 & 4.49 & 4.70 & 4.94 \\
     BioMedLM & 3.74 & 4.10 & 4.06 & 4.46 & 4.61 & 4.96 \\
     LLaMA 2 & 3.92 & 4.29 & 4.16 & 4.56 & 4.69 & 4.92 \\
     \textbf{Mistral 7B} & \textbf{4.21} & \textbf{4.45} & \textbf{4.43} & \textbf{4.66} & \textbf{4.79} & \textbf{4.97} \\
     \bottomrule
    \end{tabular}
    \caption{MultiMedQA 140 Rubric Scores (Average)}
    \label{tab:medmcqa}
\end{table}
\FloatBarrier

\begin{figure}[ht]
\centering
\includegraphics[width=0.60\textwidth]{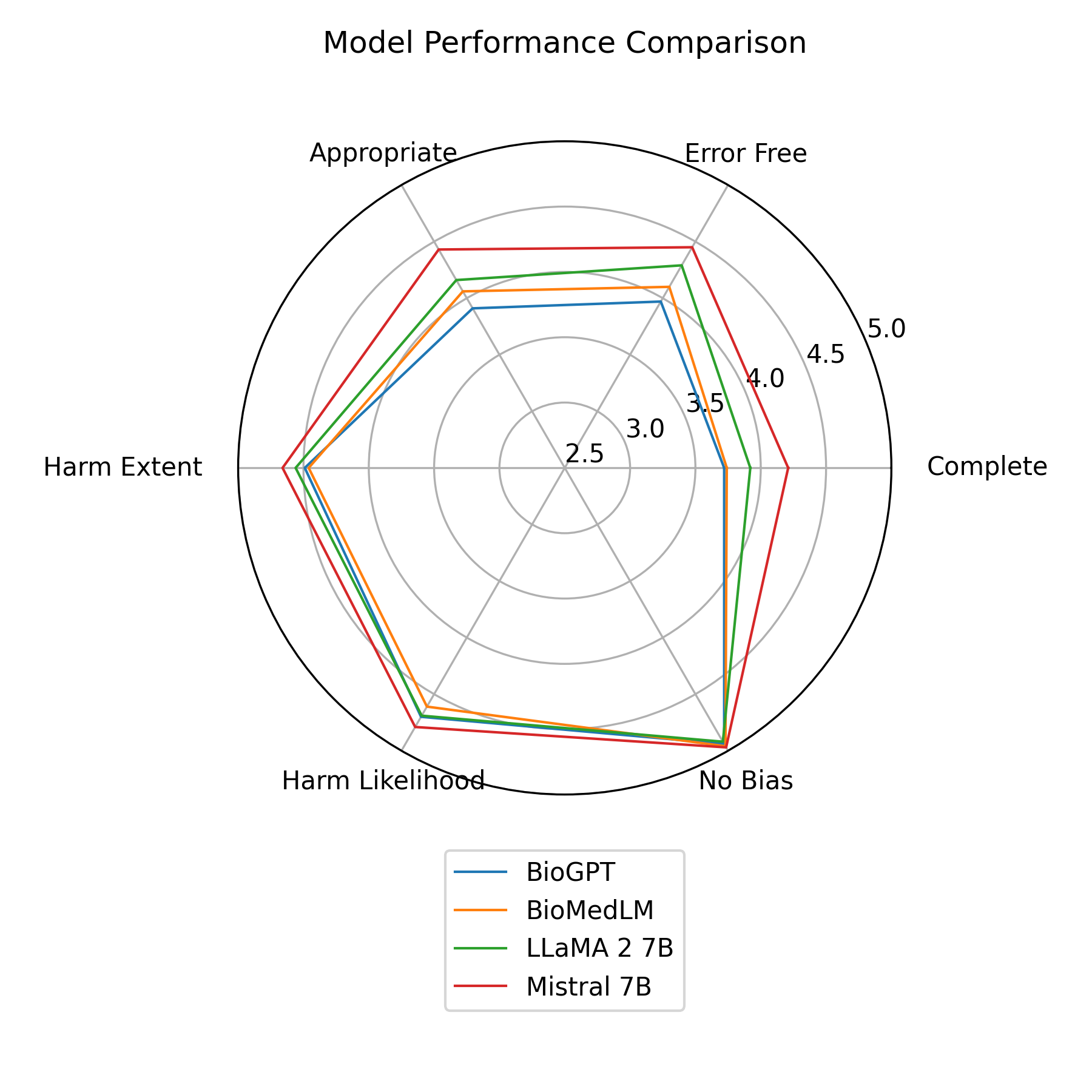}
\caption{\label{fig:consumer_qa_1} Different Model Performance Across The Six Metrics}
\end{figure}

\begin{table}[ht]
    \centering
    \begin{tabular}{l l l l l l l}
     \toprule
     Model & Complete & Error Free & Appropriate & Harm & Harm & Bias \\
      & & & & Extent & Likelihood & \\
     \midrule
     BioGPT-large vs. \\ BioMedLM & 0.85166 & 0.40673 & 0.27529 & 0.99184 & 0.50446 & 0.51209 \\
     BioGPT-large vs.  \\ LLaMA 2 7B & 0.18225 & 0.0256* & 0.10895 & 0.46234 & 0.75242 & 0.5316 \\
     BioGPT-large vs. \\ Mistral 7B & 0.0003* & 0.00052* & 0.00015* & 0.11992 & 0.18566 & 0.44453 \\
     BioMedLM vs. \\ LLaMA 2 7B & 0.25027 & 0.15724 & 0.61647 & 0.45822 & 0.34591 & 0.99293 \\
     BioMedLM vs. \\ Mistral 7B & 0.00052* & 0.00786* & 0.00778* & 0.13359 & 0.05232 & 0.93364 \\
     LLaMA 2 7B vs. \\ Mistral 7B & 0.02089* & 0.22267 & 0.02358* & 0.45358 & 0.33425 & 0.92705 \\
     \bottomrule
    \end{tabular}
    \caption{Significance level of score differences across six metrics under the Mann-Whitney U test (* denotes significance $p < 0.05$)}
    \label{tab:model_comparison_significance}
\end{table}

\bigskip
\paragraph{Performance Across All Questions} Across all metrics, Mistral 7B was the strongest performer on this task (see figure \ref{fig:consumer_qa_1}). The indicator that all models scored best on is the no bias metric, followed by harm likelihood, harm extent, error free, appropriate, and finally complete. For the complete metric, Mistral 7B, significantly (see table \ref{tab:model_comparison_significance}) outperforms the other three models, with an average score of 4.21. For the error-free and appropriate metrics, Mistral (4.45, 4.43) performs significantly better than BioGPT (3.97, 3.91) and BioMedLM (4.10, 4.06), but LLaMA 2 (4.29, 4.16) performs on par with it. For the harm extent, harm likelihood and bias indicators, there was no significant difference between all indicators, especially for bias, where the average across all models is greater than 4.90.

\begin{figure}[ht]
\centering
\includegraphics[width=1.0\textwidth]{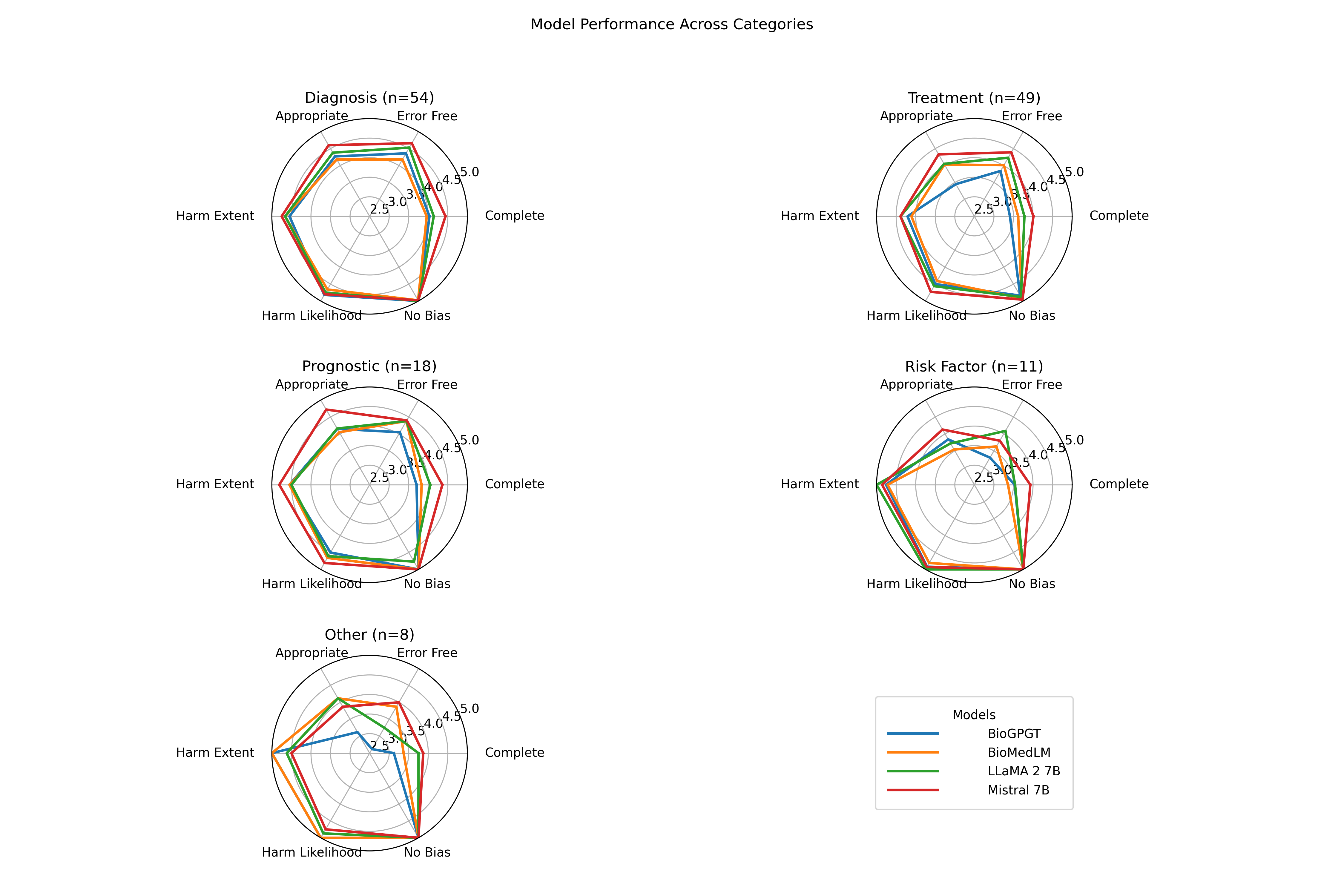}
\caption{\label{fig:consumer_qa_2} Different Model Performance Across The Six Metrics By Question Category}
\end{figure}

\paragraph{Performance By Question Category} Examining question types, for the categories of diagnosis, treatment, and prognostic, the distribution of scores is roughly that of the overall scores, with Mistral performing the best, and the larger 7B English models outperforming the smaller biomedical models (see figure \ref{fig:consumer_qa_2}). Exceptions to these trends occurred in the risk factor (n=11) and other categories (n=8), both of which were small in size. For instance in the risk factor category, LLaMA was deemed safer than Mistral and to make less errors. BioMedLM outperformed Mistral in appropriateness and the safety categories in the other category. The average scores per category are list in section C of the supplementary material.

\section{Discussion}

\subsection{MedQA Performance}

The present state of the art on this task is GPT-4 with Medprompt: 90.2\%, closely followed by Med-PaLM 2 fine-tuned on MedQA+MedMCQA and utilizing ensemble refinement: 86.5\%. For sub-10B scale models, Meditron 7B \citep{chen2023meditron70b} scores 52.0\%, and BioMedLM (with specialized architecture) scores 54.7\% \citep{bolton2024biomedlm}.

\begin{table}[ht]
    \centering
    \begin{tabular}{l r}
     \toprule
     Model & Accuracy \\
     \midrule
     BioMedLM (w/MedMCQA data + MC architecture) & 54.7 \\
     Mistral 7B (w/MedMCQA data) & 63.0 \\
     MEDITRON 70B & 70.2 \\
     Med-PaLM 2 & 86.5 \\
     GPT-4 (w/Medprompt) & 90.2 \\
     \bottomrule
    \end{tabular}
    \caption{Performance Of Various Models On MedQA}
    \label{tab:medqa-various-models}
\end{table}
\FloatBarrier

Mistral 7B’s score of 63.0\% on this benchmark approaches the score achieved by the original Med-PaLM system of 67.6\% and sets a new standard for what can be achieved for models with less than 10B parameters, though it still falls short of a passing grade and is substantially lower than scores achieved by large language models. There are multiple avenues to improving sub-10B parameter model performance on MedQA, including continued pretraining of the model on biomedical text, increasing model size, utilizing retrieval augmented generation, utilizing chain of thought, and distillation of frontier models. Starting from a score of 63.0\%, we hope future researchers can achieve a passing score with a more compact model.

The hyperparameter sweep for this experiment was important to get a fair sense of how the models compare against each other. This stands in contrast to other work that has evaluated language model performance on the MedQA task. For example Meditron-70B presents a comparison between Mistral 7B (instruct version) and LLaMA 7B on MedQA. While LLaMA 7B is fine-tuned on the MedQA training data, the authors only report the direct performance of the out-of-the-box Mistral 7B (instruct version). This table produces a misleading impression that Mistral 7B scores 41.1 vs.\ LLaMA 7B’s score of 49.6. When comparing model performance, it is crucial to keep as many elements as similar as possible.

It is important to note the limitations of this study. We provide only a small set of hyperparameter sweeps because it becomes computationally prohibitive to do larger ones. The effect of other settings such as input/output formats, sequence length, and other miscellaneous weight settings (e.g., weight decay) was not explored. Ultimately MedQA is a multiple choice test, so performance on this task is only a proxy for a model’s ability to recall medical knowledge and reason about medical scenarios. More realistic, human-evaluated benchmarks are needed to truly assess the capabilities of these models.

\subsection{Performance On MultiMedQA Long Form Question Answering}

\subsubsection{Physican Findings On MultiMedQA Long Form Question Answering}

In contrast to the slightly contrived task of MedQA, this free-response question-answering task relies on the generative capabilities of the models we evaluated.

The physician review team consistently rated Mistral 7B the highest on a variety of metrics and consistently preferred the English 7B models to the specialist models. This pattern generally held up across different question types related to diagnosis, treatment, and prognosis, highlighting Mistral's robust performance across question categories.  Amongst the remaining categories (``risk factor" n=11, and ``other" n=8), smaller sample sizes limited the ability to draw definitive conclusions about the different models. Mistral was still highly rated on many benchmarks in these categories, but there were exceptions such as LLaMA 2 being rated as producing less errors in the ``risk factor" category.

While not perfect, Mistral achieved scores that indicate reasonable responses to consumer health questions represented in the MultiMedQA dataset. That being said, the response quality of Mistral 7B is not high enough to deploy the system to production yet in a scenario where patients would directly rely on the answers of the system. For instance, it is well known language models suffer from hallucination \citep{xie2023hallucination}, producing incorrect medical information. Even the highest scoring model Mistral 7B was determined to produce over a half an error per response on average according to the physician reviewers.

It would be more realistic to use these Mistral 7B responses as the initial draft of a response to the question that a trained physician or nurse could modify and correct based on retrieved source material, for instance helping a health expert write an FAQ page by writing the first draft of the answers. We feel that the promising results of Mistral 7B with basic fine-tuning suggest it is plausible to build a production level system with a 10B--20B model. It is also important that utilization of Mistral 7B in the clinical environment is prospectively evaluated for accuracy and potential adverse impact on the receiver before concluding clinical effectiveness or applicability.

It is interesting to note that while Mistral 7B substantially outperformed models on the MedQA task, the gap between Mistral 7B and LLaMA 7B on this more realistic paragraph response task was much smaller. This suggests that multiple choice performance is not a good proxy for performance on tasks requiring paragraph responses.

\subsubsection{Physician panel discussion on questions}

There are many sources of noise in this evaluation. We only sample 140 examples from the HealthSearchQA dataset, which is a sample of all possible 3375 consumer questions. We sample three physicians from the global community, and have a random assignment of questions to physicians and the ordering of the presentation of questions.

In addition, as much as we introduce a training session for the clinicians, they are subjective, and the quality of the question dataset lends itself to ambiguity. The clinicians have different training backgrounds and specialties, meaning that their risk thresholds may also differ based on the topic of interest.  For example, a family physician may have a different risk tolerance from an interventional cardiologist. The questions themselves in the HealthSearchQA dataset also had varying qualities, with spelling mistakes and ambiguous meaning, that risked shifting a physician’s grading of the response. This problem was partially mitigated by introducing the appropriateness metric and training and agreement of the scoring physicians on a small sample initially, as a way to assess how appropriately the model responded to the heart of the question, including addressing issues of disease severity and clinical urgency posed by a question. The implications of different physician specialities, along with the vast difference between public facing and professionally curated prompts, accounts for the difficulty in this task.

Below we present physician commentary about the HealthSearchQA task questions.

\paragraph{Flawed Questions}

The questions in the MultiMedQA 140 dataset are derived from layperson questions from internet (Google) search terms. Coming from non-clinicians and non-scientists, some questions reflected incorrect assumptions or an incomplete understanding of clinical medicine which in turn influenced the quality of answers generated by the algorithm. For example, consider the question: “What are the three types of angina?” In reality there are many ways to categorize angina, and there are no predominant “3 types” of angina used within routine clinical practice. As such this question is inherently flawed, with the result that the large language model must generate an answer to a question that is not based on the reality of medical knowledge or norms of categorization, thereby setting up the model to perform poorly in its response.

\paragraph{Ambiguous Questions}

A proportion of questions were ambiguous in their meaning. Ambiguity may cause problems where a question has crucial information missing, requiring the model to make significant assumptions. For example, the question “What medication should a 65 year old male use for leg pain?” appears simple enough, however a high quality response to this question first requires us to know the underlying disease causing the pain, the duration of this problem, the patient’s medical history and current medications at a bare minimum. Different medications, comorbidities and other factors such as demographic, age might lead to major differences in the advice dispensed within the model response. Yet without this information provided as a complete dataset in the prompt,  the model is in a perilous position, attempting to fill in the gaps in a way that is medically sound, valid and safe.

\paragraph{Question Complexity And Hallucination}

A general finding is that the more complex the question, the greater variability in the generated response. While straightforward questions like “What defines obese?” generally garnered similar responses, complex, multi-part questions such as “Is 50,000 IUs per week of Vitamin D safe and can it cause flatulence?” steered models into poor responses with hallucination.

In one of the responses (see table \ref{tab:response-w-hallucination}) to the Vitamin D question, “fecal incontinence” is mentioned in the response, which is erroneous. Furthermore, the model clearly attempts to offer guidance in the situation of massive fecal incontinence by adding that one may be advised to take “electrolyte supplementation” if this were to be the case. When taken in isolation, this advice is correct. However it appears that the conclusion of the answer is based upon an inappropriate amalgamation of “Does vitamin D supplementation cause flatulence?” and “what is the management of massive fecal incontinence or diarrhea?”. Model hallucination is a significant limitation of health models and there must be a way of regulating/assessing this moving forwards There are instances where the information given may be correct independently, but its interpretation is completely irrelevant and false in the context of the question.

\begin{table*}[h]
    \centering
    \begin{tabular}{p{15cm}}
    \toprule
    \bigskip
    Question: \\
    \bigskip
    Is 50,000 IUs per week of Vitamin D safe and can it cause flatulence? \\
    \bigskip

    Answer: \\
    \bigskip
    
    You may be at risk for vitamin D toxicity if you take higher than prescribed doses of vitamin D. Your provider may need to measure your
    blood levels of vitamin D or you may be at risk for vitamin D deficiency. Vitamin D toxicity and vitamin D deficiency can have the same signs and symptoms. Also, if you take 50,000 IUs of vitamin D daily, you may notice increased stool volume (fecal incontinence). Your provider may suggest an electrolyte supplement. 
    \bigskip \\

    \bottomrule
    \end{tabular}
    \caption{Example Response With Hallucination}\label{tab:response-w-hallucination}
\end{table*}

\subsubsection{Mistral 7B Responses And Comparison To Med-PaLM 2}

The fine-tuned Mistral 7B consistently scored above four across each dimension of analysis, suggesting it can often produce answers that generally cover the needed material for a quality response, while avoiding medical errors (averaging 0.55 errors per response) and maintaining safety.
Specific examples of the best Mistral 7B responses can be found in table \ref{tab:consumerqa-mistral-best-responses}.

\begin{table*}[htbp]
    \centering
    \small
    \begin{tabular}{p{3.75cm} p{11cm}}
     \toprule
     Question & Mistral 7B Answer \\
     \midrule
     How do you know if you have herpetic whitlow? & {The signs of a herpetic whitlow include blisters and swelling in and around your fingertip. They’re usually painful. The symptoms appear suddenly and last up to two to four weeks. They start as small blisters and progress to crusty lesions.
You should see a healthcare provider if you experience any of these symptoms, especially if your blisters are painful or have infected the area around them.}\\
    \midrule
    What causes you to get ringworm? & {Fungi are a type of organism. You have fungi on your skin all the time. The most common type of ringworm is a fungal infection caused by a type of fungi called a dermatophyte. The fungi enter your body through a small cut or scratch in your skin. The infection may spread to other parts of your body.
Ringworm can also spread through contact with an infected person or animal. The fungi can live for up to three weeks on things like clothing, towels and bedding.} \\
    \midrule
    What is RSI repetitive strain injury? & {Repetitive strain injury (RSI) is a term for any of several medical conditions associated with the performance of repetitive tasks that may be associated with sustained or intense muscle activity. The affected tissues may include muscle, tendon, or nerve tissues. RSI does not necessarily result from overuse, although some types of RSIs may be associated with overuse. It also does not always result from repeated exertion at a single job, though this is frequently a contributing factor. There is no consensus on the definition of repetitive strain injury.} \\
    \midrule
    Is PCOS and hirsutism the same? &
    {Hirsutism is a common symptom of polycystic ovary syndrome (PCOS). People with hirsutism have excess hair growth in a pattern typically seen in men. PCOS is an endocrine disorder that affects how the ovaries function. People with PCOS also have high androgen levels, which may cause hirsutism.} \\
\midrule
What condition causes a reduced sense of touch? & {Diabetic neuropathy is the most common condition that causes a reduced sense of touch. If you have this condition, you may have tingling, numbness or burning sensations in your feet. Over time, you may lose the ability to feel heat, cold and pain in your feet. 

\bigskip
Other conditions that can cause a reduced sense of touch include:
\bigskip

Vitamin B12 deficiency.

Alcohol use disorder.

Autoimmune diseases.

Hypothyroidism.

Guillain-Barré syndrome.

HIV.

Spinal cord injury.

Leprosy.

Drug use.\bigskip} \\
    \bottomrule
    \end{tabular}
    \caption{Mistral 7B Responses To Consumer Health Questions}\label{tab:consumerqa-mistral-best-responses}
\end{table*}

The  scale we used was more detailed to produce fine-grained assessments, whereas the Med-PaLM 2 team used binary questions which lead to softer conclusions. While the rubric used by our physicians does not perfectly map to the Med-PaLM 2 rubric, it is possible to make some comparisons. For instance, a score of five on the “no bias” category can be mapped to the binary score of not having bias using the Med-PaLM 2 review criteria.

Along some dimensions of review, Mistral 7B’s responses had comparable quality to those produced by Med-PaLM 2. Mistral 7B scored 98.4\% on having no bias vs.\ Med-PaLM 2’s 97.1\%. Our physicians rated Mistral 7B as being unlikely to cause harm 97.1\% of the time vs.\ Med-PaLM 2’s reviewers rating the harm likelihood ``low'' 95.5\% of the time. 97.9\% of Mistral 7B responses were given at least a three on both complete and appropriate which could be construed to mean a response showed evidence of question comprehension. Only 97.1\% of Med-PaLM responses were deemed to show evidence of question comprehension.

But many of the dimensions are more favorable to Med-PaLM 2. Mistral 7B only produced responses rated as incapable of producing harm 85.7\% of the time, while physicians said Med-PaLM 2’s responses were completely safe 93.3\% of the time. Med-PaLM 2 showed much more ability to avoid error than Mistral 7B. While Mistral 7B only scored a perfect 5 on avoiding medical error 73.1\% of the time, Med-PaLM 2 never scored below 90\% on any of the categories related to error.

\subsection{Future directions}

There are many future improvements that could lead to a system that rivals the GPT-4 and Med-PaLM 2 systems with a much smaller model size. We could use larger and higher quality foundation models with more biomedical focus in the 10B--20B scale. A larger and higher quality question-answer training set would reduce variance, especially if it was curated by medical professionals, rather than targeted towards consumers. Augmentation with retrieval results, R.A.G., is particularly salient in a knowledge-intensive field like medicine \citep{lewis2021retrievalaugmented}. Additional models such as rules-based processes for face checking against a knowledge base, reinforcement learning (RLHF) \citep{ouyang2022training} or direct preference optimization (DPO) \citep{rafailov2023direct}. These directions provide exciting possibilities for future evaluation on large language models in biomedicine.

\section{Conclusion}
Mistral 7B performed the best on both tasks evaluated. BioMedLM is a compromise that is smaller than the 7B models but can still perform reasonably well. BioGPT-large can produce acceptable results if one lacks the compute for using the larger models. On both tasks, we found that the larger scale models trained on general English (which may have included the PubMed corpus) outperformed the smaller domain-specific models. It is unclear if a scaled up biomedical specialist model would offer much of an improvement over Mistral 7B. Before models from this class can be applied in a clinical setting, it is important that outputs are reviewed by medical experts.

\newpage
\bibliographystyle{plainnat}
\bibliography{references}

\clearpage
\appendix
\counterwithin{figure}{section}
\counterwithin{table}{section}
\input{appendix}

\end{document}

%% file: appendix.tex
\renewcommand{\thesubsection}{\arabic{subsection}}

\section{Example Questions}
\setcounter{subsection}{0}

\subsection{MultiMedQA Long Form Questions}

\begin{table*}[ht]
    \centering
    \begin{tabular}{p{8cm} p{7.2cm}}
    \toprule
    Question \\
    \midrule
    How long does prickly heat rash last? \\
    
    What happens if listeria is left untreated? \\
    
    Who is at risk for molar pregnancy? \\
    
    Is mumps contagious yes or no? \\
    
    What is maximum daily dosage of Metformin? \\
    \bottomrule
    \end{tabular}
    \caption{Example Questions For MultiMedQA Long Form Response}\label{tab:multimedqa-examples}
\end{table*}
\FloatBarrier

\bigskip
\bigskip
\begin{table*}[ht]
    \centering
    \begin{tabular}{p{15cm}}
    \toprule
    \bigskip
    Question: \\
    \bigskip
    
    Text of question... \\
    \bigskip
    
    \textbf{Answer:} \\
    \bigskip

    \textbf{Text of answer...}
    
    \bigskip
    
    \\ \bottomrule 
    \end{tabular}
    \caption{MultiMedQA 140 Input/Output Format (expected output in bold)}\label{tab:multimedqa-format}
\end{table*}
\FloatBarrier

\newpage
\subsection{MedQA}

\bigskip
\begin{table*}[ht]
    \centering
    \begin{tabular}{p{9cm} p{6cm}}
    \toprule
    Question & BioMedLM Answer \\
    \midrule
    A 45-year-old woman presents to the emergency department with acute onset of severe right upper quadrant abdominal pain that radiates to the infrascapular region. Her medical history is significant for obesity, hypertension, obstructive sleep apnea, and gastric bypass surgery 2 years ago after which she lost 79 kg (150 lb). The patient complains of nausea and vomiting that accompanies the pain. Her temperature is 38.9°C (101.2°F), blood pressure is 144/88 mm Hg, heart rate is 76/min, and respiratory rate is 14/min (fever). Abdominal examination is significant for right upper quadrant tenderness along with guarding and cessation of inspired breath on deep palpation of the right upper quadrant. Which test should be ordered first for this patient? & {\textbf{A) Abdominal ultrasound}
    
    B) CT scan of the abdomen
    
    C) MRI of the abdomen
    
    D) X-ray film of the abdomen} \\
    \bottomrule
    \end{tabular}
    \caption{Example Question for MedQA}\label{tab:medqa-examples}
    \end{table*}
    \FloatBarrier

\bigskip
\bigskip
\begin{table*}[ht]
    \centering
    \begin{tabular}{p{15cm}}
    \toprule
    \bigskip
    Question: \\
    \bigskip
    
    Text of question \\
    \bigskip
    
    Options: \\
    \bigskip
    
    A) ...
    
    B) ...
    
    C) ...
    
    D) ...
    
    \bigskip
    
    \textbf{Answer: A}
    
    \bigskip
    
    \\ \bottomrule
    \end{tabular}
    \caption{MedQA Input/Output Format (expected output in bold)}\label{tab:medqa-format}
\end{table*}
\FloatBarrier

\bigskip
\bigskip
\begin{table*}[ht]
    \centering
    \begin{tabular}{p{15cm}}
    \toprule
    \bigskip
    You are a medical doctor taking the US Medical Licensing Examination. You need to demonstrate your understanding of basic and clinical science, medical knowledge, and mechanisms underlying health, disease, patient care, and modes of therapy. Show your ability to apply the knowledge essential for medical practice. For the following multiple-choice question, select one correct answer from A to E. Base your answer on the current and standard practices referenced in medical guidelines. \\

    \bigskip
    Question: \\
    \bigskip
    
    Text of question \\
    \bigskip
    
    Options: \\
    \bigskip
    
    A) ...
    
    B) ...
    
    C) ...
    
    D) ...
    
    \bigskip
    
    \textbf{Answer: A) Text Of Answer}
    
    \bigskip
    
    \\ \bottomrule
    \end{tabular}
    \caption{MedQA Input/Output Format For Additional Mistral 7B experiments (expected output in bold)}\label{tab:medqa-format-mistral}
\end{table*}
\FloatBarrier

The introductory text in the prompt is adapted from \citep{chen2023meditron70b}

\clearpage
\section{Hyperparameter Sweep Details}

\begin{table}[ht]
    \centering
    \begin{tabular}{l r l r}
     \toprule
     Model & Learning Rate & Batch Size & Epochs\\
     \midrule
     BioGPT-large & {5e-06, 1e-05, 2e-05} & {16} & {3,5,10,20} \\
     BioMedLM & {1e-06, 2e-06, 5e-06} & {16} & {3,5,10,20} \\
     LLaMA 2 & {5e-07, 1e-06, 2e-06} & {16} & {3,5,10,20} \\
     Mistral 7B & {5e-07, 1e-06, 2e-06} & {16} & {3,5,10,20} \\
     \bottomrule
    \end{tabular}
    \caption{MedQA Hyperparameter Sweep Settings}
    \label{tab:medqa-hyperparams}
\end{table}

\bigskip

\begin{table}[ht]
    \centering
    \begin{tabular}{l r l r}
     \toprule
     Model & Learning Rate & Batch Size & Epochs\\
     \midrule
     Mistral 7B & {4e-07, 5e-07, 1e-06, 2e-06} & {16} & {1,2,3,5} \\
     \bottomrule
    \end{tabular}
    \caption{Mistral 7B Extra Training w/MedMCQA Training Data Hyperparameter Sweep Settings}
    \label{tab:medmcqa-mistral}
\end{table}

\bigskip

\begin{table}[ht]
    \centering
    \begin{tabular}{l r l r}
     \toprule
     Model & Learning Rate & Batch Size & Epochs\\
     \midrule
     BioGPT-large & {5e-06, 1e-05, 2e-05} & {16} & {1,2,5} \\
     BioMedLM & {1e-06, 2e-06, 5e-06} & {16} & {1,2,5} \\
     LLaMA 2 & {5e-07, 1e-06, 2e-06} & {16} & {1,2,5} \\
     Mistral 7B & {5e-07, 1e-06, 2e-06} & {16} & {1,2,5} \\
     \bottomrule
    \end{tabular}
    \caption{MultiMedQA 140 Hyperparameter Sweep Settings}
    \label{tab:multimedqa}
\end{table}

\clearpage
\section{Physician Rating By Category}

\begin{table}[htbp!]
    \centering
    \begin{tabular}{l l l l l l l}
     \toprule
     Model & Complete & Error Free & Appropriate & Harm  & Harm & Bias \\
      & & & & Extent & Likelihood & \\
     \midrule
     BioGPT-large & 4.03 & 4.36 & 4.27 & 4.55 & 4.82 & 5.00 \\
     BioMedLM & 3.96 & 4.18 & 4.18 & 4.66 & 4.66 & 4.98 \\
     LLaMA 2 & 4.14 & 4.53 & 4.38 & 4.66 & 4.75 & 5.00 \\
     Mistral 7B & 4.44 & 4.66 & 4.60 & 4.75 & 4.80 & 4.98 \\
     \bottomrule
    \end{tabular}
    \caption{MultiMedQA 140 Rubric Scores (Diagnosis n=54)}
    \label{tab:consumerqa-diagnostic}
\end{table}

\begin{table}[htbp!]
    \centering
    \begin{tabular}{l l l l l l l}
     \toprule
     Model & Complete & Error Free & Appropriate & Harm  & Harm & Bias \\
      & & & & Extent & Likelihood & \\
     \midrule
     BioGPT-large & 3.41 & 3.84 & 3.45 & 4.21 & 4.50 & 4.84 \\
     BioMedLM & 3.62 & 4.01 & 4.03 & 4.11 & 4.41 & 4.92 \\
     LLaMA 2 & 3.78 & 4.23 & 4.05 & 4.39 & 4.56 & 4.88 \\
     Mistral 7B & 4.01 & 4.39 & 4.33 & 4.39 & 4.73 & 4.96 \\
     \bottomrule
    \end{tabular}
    \caption{MultiMedQA 140 Rubric Scores (Treatment n=49)}
    \label{tab:consumerqa-treatment}
\end{table}

\begin{table}[htbp!]
    \centering
    \begin{tabular}{l l l l l l l}
     \toprule
     Model & Complete & Error Free & Appropriate & Harm  & Harm & Bias \\
      & & & & Extent & Likelihood & \\
     \midrule
     BioGPT-large & 3.70 & 4.05 & 4.15 & 4.55 & 4.50 & 5.00 \\
     BioMedLM & 3.83 & 4.38 & 4.05 & 4.55 & 4.66 & 5.00 \\
     LLaMA 2 & 4.05 & 4.38 & 4.16 & 4.50 & 4.61 & 4.77 \\
     Mistral 7B & 4.36 & 4.40 & 4.72 & 4.81 & 4.81 & 5.00 \\
     \bottomrule
    \end{tabular}
    \caption{MultiMedQA 140 Rubric Scores (Prognostic n=18)}
    \label{tab:consumerqa-prognostic}
\end{table}

\begin{table}[htbp!]
    \centering
    \begin{tabular}{l l l l l l l}
     \toprule
     Model & Complete & Error Free & Appropriate & Harm  & Harm & Bias \\
      & & & & Extent & Likelihood & \\
     \midrule
     BioGPT-large & 3.53 & 3.30 & 3.84 & 4.76 & 4.92 & 5.00 \\
     BioMedLM & 3.36 & 3.63 & 3.54 & 4.72 & 4.81 & 5.00 \\
     LLaMA 2 & 3.54 & 4.09 & 3.72 & 5.00 & 5.00 & 5.00 \\
     Mistral 7B & 3.93 & 3.80 & 4.13 & 4.86 & 4.93 & 5.00 \\
     \bottomrule
    \end{tabular}
    \caption{MultiMedQA 140 Rubric Scores (Risk factor n=11)}
    \label{tab:consumerqa-risk-factor}
\end{table}

\clearpage
\begin{table}[htbp!]
    \centering
    \begin{tabular}{l l l l l l l}
     \toprule
     Model & Complete & Error Free & Appropriate & Harm  & Harm & Bias \\
      & & & & Extent & Likelihood & \\
     \midrule
     BioGPT-large & 3.12 & 2.62 & 3.12 & 5.00 & 5.00 & 5.00 \\
     BioMedLM & 3.37 & 3.87 & 4.12 & 5.00 & 5.00 & 5.00 \\
     LLaMA 2 & 3.75 & 3.25 & 4.12 & 4.62 & 4.87 & 5.00 \\
     Mistral 7B & 3.87 & 4.00 & 3.87 & 4.50 & 4.75 & 5.00 \\
     \bottomrule
    \end{tabular}
    \caption{MultiMedQA 140 Rubric Scores (Other n=8)}
    \label{tab:consumerqa-other}
\end{table}